\documentclass{article}

\usepackage[english]{babel}

\usepackage[letterpaper,top=2cm,bottom=2cm,left=3cm,right=3cm,marginparwidth=1.75cm]{geometry}

\usepackage{amsmath}
\usepackage{graphicx}
\usepackage[colorlinks=true, allcolors=blue]{hyperref}
\usepackage{cite}
\usepackage[sort,numbers]{natbib}
\usepackage{booktabs}
\usepackage{multicol}
\usepackage{subfigure}
\usepackage{bbding}

\usepackage{authblk}

\title{\LARGE{\textbf{(Fusionformer):Exploiting the Joint Motion Synergy with Fusion Network Based On Transformer for 3D Human Pose Estimation}}}
\author[*]{Xinwei Yu}
\author[**]{Xiaohua Zhang}
\affil[*]{Hangzhou Research Institute, Xidian University}
\affil[**]{Xidian University}
\date{}
\begin{document}
\maketitle

\textbf{Abstract.}For the current 3D human pose estimation task, a group of methods mainly learn the rules of 2D-3D projection from spatial and temporal correlation. However, earlier methods model the global features of the entire body joint in the time domain, but ignore the motion trajectory of individual joint. The recent work \cite{zhang2022mixste} considers that there are differences in motion between different joints and deals with the temporal relationship of each joint separately. However, we found that different joints show the same movement trends under some specific actions. Therefore, our proposed  \textbf{Fusionformer} method introduces a self-trajectory module and a mutual-trajectory module based on the spatio-temporal module .After that, the global spatio-temporal features and local joint trajectory features are fused through a linear network in a parallel manner. To eliminate the influence of bad 2D poses on 3D projections, finally we also introduce a pose refinement network to balance the consistency of 3D projections. In addition, we evaluate the proposed method on two benchmark datasets (Human3.6M, MPI-INF-3DHP). Comparing our method with the baseline method poseformer,the results show an improvement of 2.4\% MPJPE and 4.3\% P-MPJPE on the Human3.6M dataset, respectively.
{\bf\emph{ Key words-\ 3D human pose estimation; joint trajectory features; Fusionformer; refinement network}\rm}

\section{Introduction}
In recent years, human pose estimation has played an increasingly important role in the field of computer vision, due to the development needs of virtual reality, motion capture, and motion recognition. 3D human pose estimation can be defined as predicting the joint positions of the human body from pictures or videos.According to previous work, there are two principal solutions to this task: (1) single-stage: end-to-end model (2) two-stage:2D lifting to 3D model. Since the first scheme directly regresses to the 3D coordinates, the source of its error cannot be identified, so the related work \cite{Pavlakos_2018_CVPR,moon2020i2l} does not produce the desired effect. Our work follows \cite{cai2019exploiting,chen2021anatomy,liu2020attention,martinez2017simple,zhang2022mixste} to train a high precision 2D lifting to 3D model. Although many excellent work \cite{zheng20213d,li2022mhformer,zhang2022mixste} has made huge progress on this task, there are still some problems. At first, the projection of 2D key points into 3D poses is a multi-solution problem, so there is a certain depth ambiguity.  In addition, some joints are occluded when performing some complex movements.
\begin{figure}[h]
\centering
\includegraphics[width=0.5\textwidth]{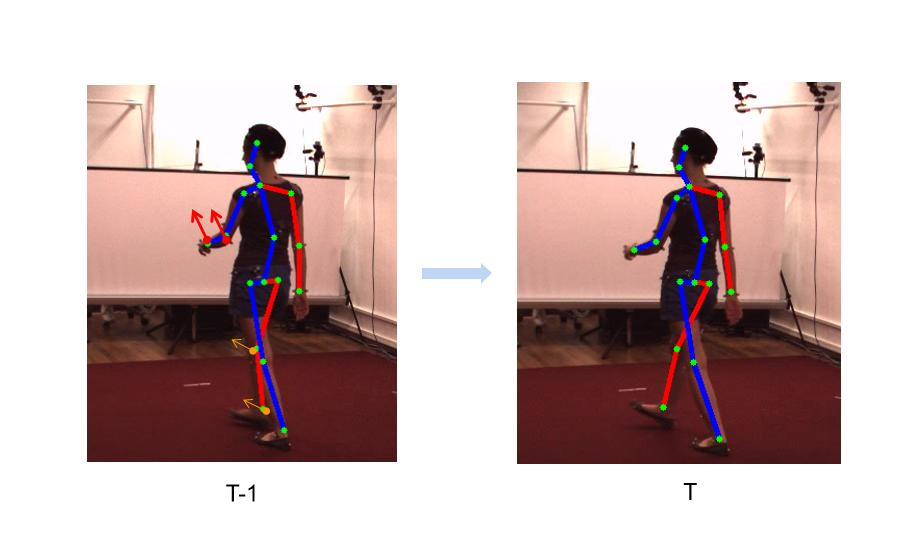}
\caption{\label{fig1}The above two pictures represent the human poses of the two adjacent frames.The arrows depict the direction of joint movement.}
\end{figure}

\begin{figure}[htbp]
\centering
\subfigure[Separate Temporal Correlation
of each joint]{
\begin{minipage}[t]{0.5\linewidth}
\centering
\includegraphics[width=1\textwidth]{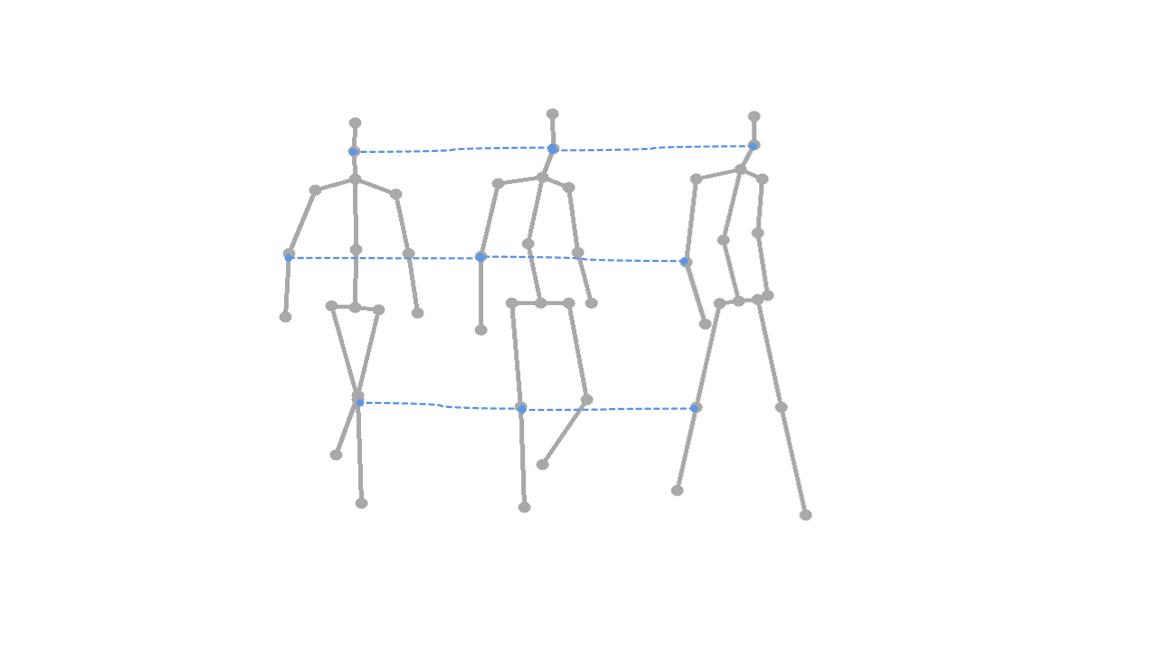}
\end{minipage}%
}%
\subfigure[Cross Temporal Correlation of different joints]{
\begin{minipage}[t]{0.5\linewidth}
\centering
\includegraphics[width=1\textwidth]{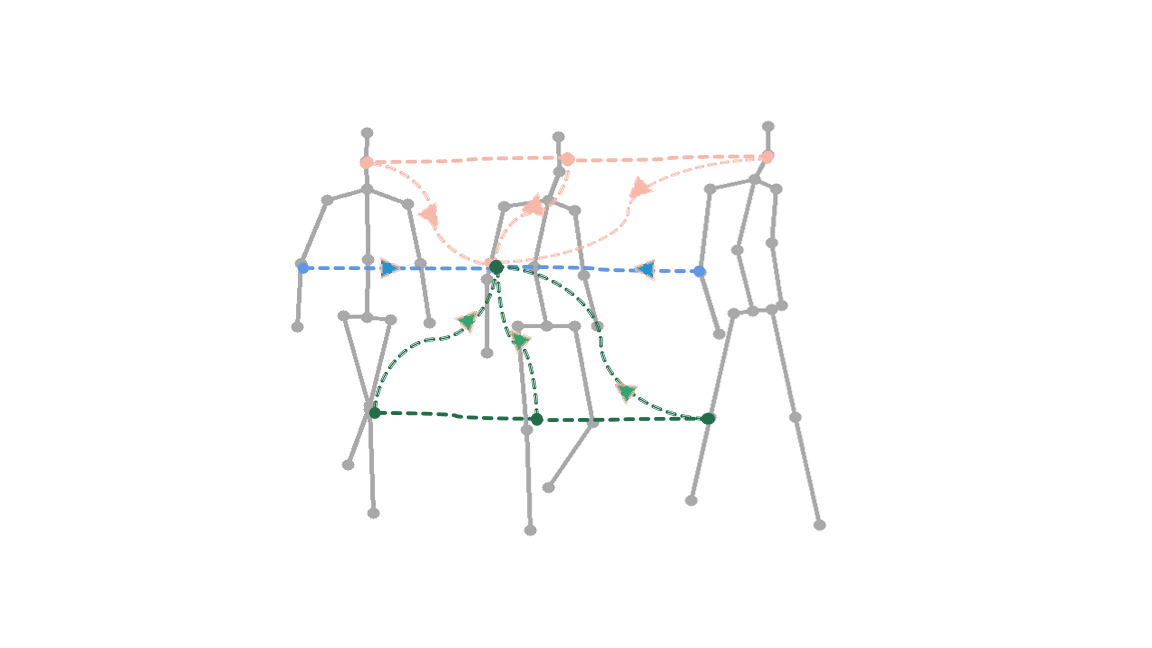}
\end{minipage}
}%

\centering
\caption{\label{fig2}(a) The left only considers the motion of the independent joints in time.(b) The right additionally adds motion prediction for different joints in adjacent frames.}
\end{figure}

\noindent In order to solve the above problems, many methods not only capture the constraint relationship between joints in the spatial domain, but also combine multiple frames of information in the time domain to obtain more robust and stable expressions. \cite{pavllo20193d} used convolutional neural network (CNN)to efficiently process information on time stream, the disadvantage is that the receptive field is limited. \cite{zheng20213d} et al. use a Transformer Encoder-based network to simultaneously model the relationship between human pose sequences in time and space. \cite{cai2019exploiting,Hu2021ConditionalDG} employ a graph convolutional network to model the spatiotemporal representation between human joints in graph form.However, they feed entire body joints into the temporal model, ignoring the predictive relationship of the local joint in the time domain.\\
A recent work \cite{zhang2022mixste} begins to focus on the temporal expression of motion in each joint (see Figure \ref{fig2} (a)). It separates all the joints and then models the relationship of each joint in the time domain through  network in a parallel manner.This method not only increases the number of network parameters to a certain extent, but also ignores the dependencies of different joints in time.Similar to mining the spatial relationship between joints, our method attempts to place all the corresponding points  in the same dimension, and then send them to the network uniformly, avoiding the redundancy of network modules(see Figure \ref{fig4} (a)).\\
As shown in Figure \ref{fig1}, we can significantly observe that the forearm at the T-moment is lifted up some distance relative to the T-1 moment. The wrist and elbow joints are all moving in the same direction. Different from \cite{zhang2022mixste}, We believe that in some joint-dependent movements, there may also be a temporal connection between adjacent joints(see Figure \ref{fig2}(b)).Since different joints can also have certain prediction relationship in time, we can model the trajectory relationship of different joints in time on this basis.And because the deep network is not easy to train, we try to improve the convergence speed of the network by parallelizing the shallow network.\\
In this paper, we propose a new network structure \textbf{Fusionformer} for 3D human pose estimation.We have introduced two branches in FusionFormer, namely: (i)Global Information Interaction Module(GIM): exploiting global features in space and time, (ii) Local Information Interaction Module(LIM): locally capturing  differences of a single joint and the synergy of different joints in temporal domain. Then, the different features output by the shallow network are connected to obtain global-local fusion features.Our method achieves excellent results on low input video frames by means of parellel-fusion.\\
Our \textbf{contributions} are three-fold as follows:
\begin{itemize}
\item For the 3D human pose, our proposed FusionFormer uses a parallel shallow network design to mine global and local information and fuse them efficiently to speed up the training and convergence of the network.  
\item By exploiting the association of different joint trajectories, the joint motion synergy of the human body is further mined, and the joint trajectories are extracted by a simple dimensional transformation method.
\item Compared to other methods, our network achieves state-of-the-art performance on both datasets at low input video frames.
\end{itemize}

\section{Related Work}
The task of 3D human pose estimation can be performed from a single perspective or multiple perspectives. However, in reality, it is difficult to obtain multi-view data, so training a single-view model has better robustness and generalization ability. The existing methods mainly adopt the two-stage 2D to 3D method, due to its good performance. We also follow the baseline method described above.\\
Earlier methods \cite{cheng2019occlusion} , manually annotated each bone. With the advent of deep neural networks, \cite{chen2021anatomy} employs stacked linear layers to regress 3D poses. \cite{pavllo20193d} employs causal convolution and free convolution to predict intermediate frames by capturing the relationship between multi-frame inputs. \cite{8626436} proposes to use recurrent networks to mine temporal relationships between inputs through long and short-term memory units. \cite{zou2021modulated} models the human pose joints in the form of a graph structure.\\
Recently, the emergence of the Bert model marks the successful application of Transformer in Natural Language Processing (NLP). After that,Vision\;Transformer based on Bert\;Encoder was also inspired to apply to the field of computer vision. \cite{zheng20213d} et al. applied a single Transformer network for the first time to discover the relationship between human pose sequences in time and space. \cite{2021Exploiting} believes that there is redundancy between the input multi-frame sequences, and the efficiency of the model is improved by removing redundancy. \cite{2022CrossFormer} pays more attention to the relationship between local joints by combining convolutional neural network (CNN) with Transformer. \cite{zhang2022mixste} Mines the trajectory of a single joint in time by separating the joints. However, separating joints may ignore the temporal relationship of different joints, which limits the model's mining of local features.\\
Therefore, our method pays more attention to the trajectory relationship of different joints in time, so as to fully preserve the local features of human poses on joints

\begin{figure}
\centering
\includegraphics[width=0.8\textwidth]{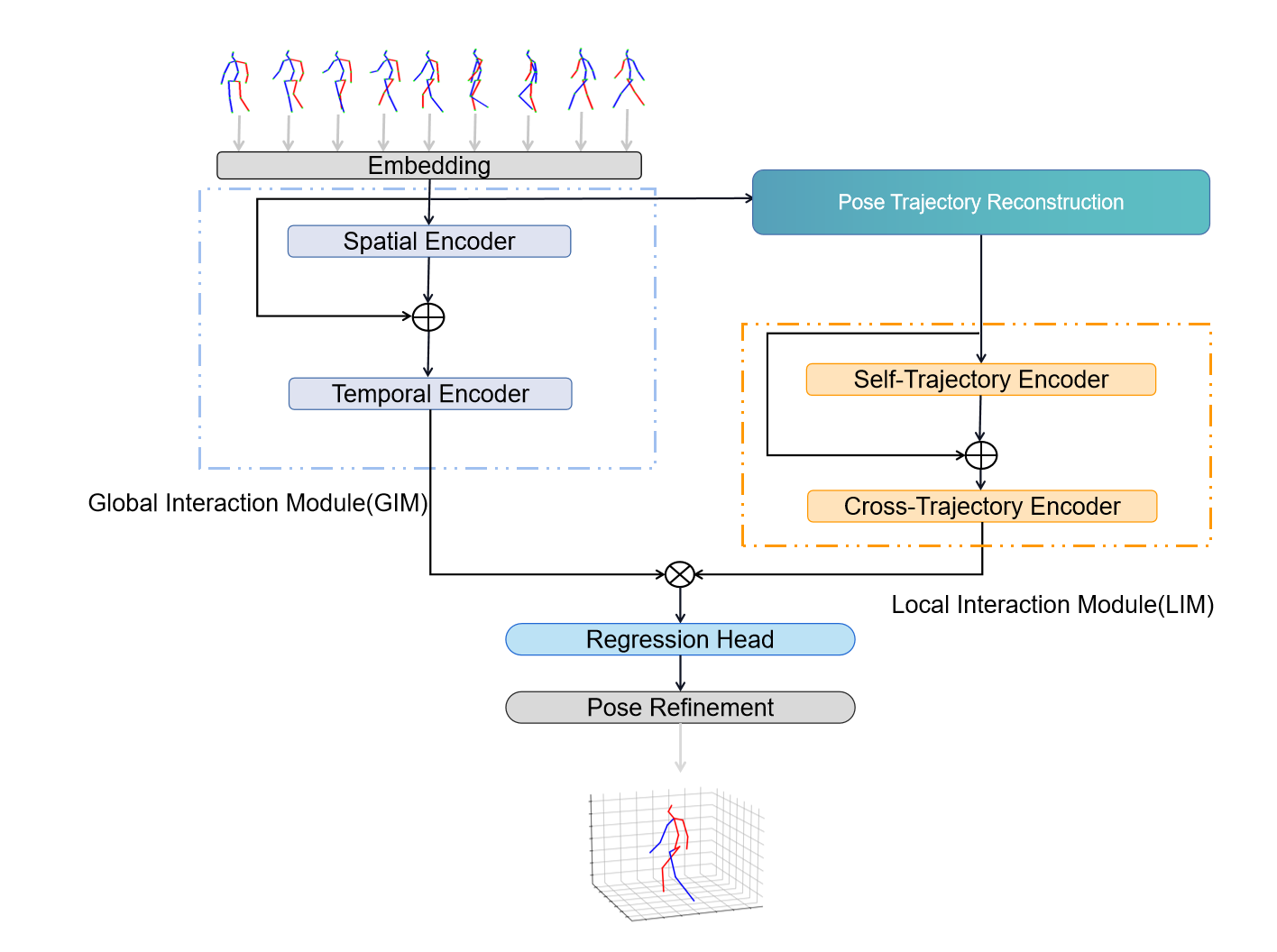}
\caption{Overview of the proposed Fusionformer}
\label{fig3}
\end{figure}

\section{Method}
In this section, we mainly describe the basic structure of the improved network. Due to the excellent performance of off-the-shelf 2D extractors, we still use the 2D-to-3D baseline method.A high-precision 2D extractor extracts the 2D pose sequence from the picture or video, and a deep neural network is then used to predict the 3D posture sequence.Due to the limitations of LSTM and RNN, we employ a Transformer-based network to predict 3D posture sequences for time series data.From a primary perspective, our architecture utilizes two parallel multi-layer network architectures, as seen in figure \ref{fig3}.The global information interaction module (GIM) on the left can extract the global features of human pose sequences in space and time.Conversely, the local information interaction module (LIM) on the other side can dig deeper into the local features between joints.The outputs of both networks are connected at the end.Such a design can separate and parallelize global features and local features, and at the same time make local features and global features fuse efficiently, avoiding the network being too deep to ignore global features.Compared with other methods, our improved network has excellent performance on 3D human pose estimation.

\subsection{Fusion Transformer}
Figure \ref{fig3} depicts the basic structure of our proposed method. The network has two main branches: a global information exchange module and a local information exchange module.First, we embed the two-dimensional human pose sequence $X=\left\{x_1,x_2,...,x_\tau,...,x_T\right\}$ ,$x_\tau\in R^{J\times D}$through the linear projection layer to obtain a high-dimensional feature vector $X_\phi\in R^{T\times{J\times{D}}}$,where T represents the number of input video frames, J represents the number of human key points in each frame of input, and 2 represents the channel of the key points.After this, we feed the high-dimensional pose sequence into the global encoder.This module contains two sub-modules respectively: spatial encoder module and temporal encoder module.The temporal encoder determines the temporal relationship between the input key points, while the spatial encoder is primarily in charge of extracting the global spatial relationship between all human key points on each frame.Then, we employ a straightforward technique to recreate the human pose, transforming the initial pose sequence into $\tilde{X}=\left\{x_1,x_2,x_3,\cdots,x_J\right\},x_j\in R^{T\times{D}}$.The reconstructed sequence models the temporal local characteristics of the joints over long distances using a local encoder.Finally, we use the regression head to connect the global pose features to obtain high-dimensional features $X_{concat}\in R^{2T\times{J\times D}}$,where D represents the embedding dimension,And the output sequence is obtained by reducing the dimension of high-dimensional features $Y\in R^{T\times{J\times3}}$.\\
\begin{figure}
\centering
\includegraphics[width=0.8\textwidth]{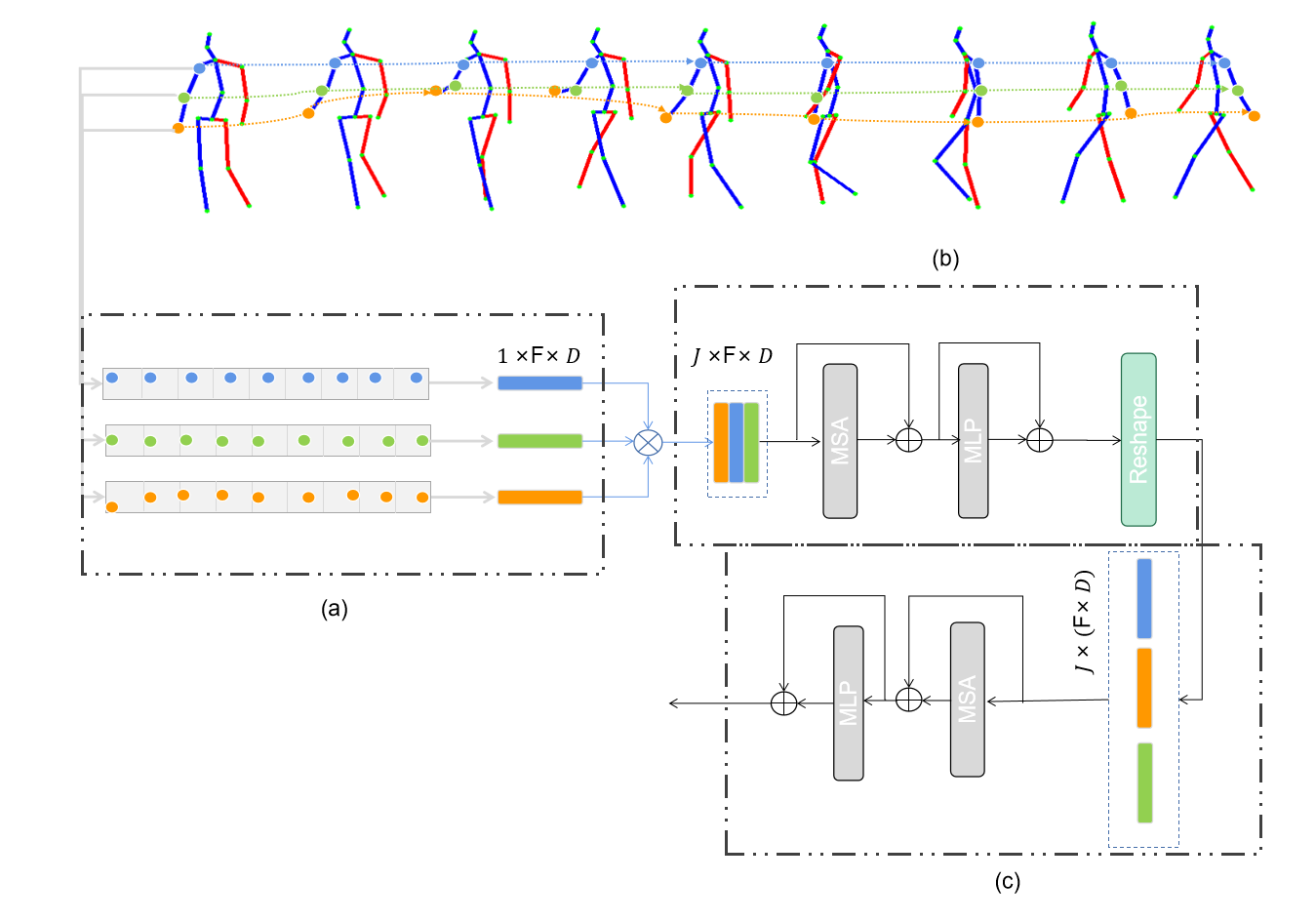} 
\caption{\label{fig4}(a) Details of Pose Trajectory Reconstruction. (b) The Network of Self-Trajectory Encoder (STE).(c) The Network of Cross-Trajectory Encoder (CTE).}
\end{figure}
\subsection{Global Interaction Module}
This section will elaborate on the details of the Global Information Interaction Module (GIM).The network structure of the global information Interaction module is shown on the left of Figure \ref{fig3}.Assuming that the number of input 2D keypoints is J, we treat each keypoint as a token.Then, each token extracts spatial depth features by embedding into high dimensions,and use learnable parameters $E_S\in R^{N\times{D}}$ to encode the spatial position of each token.After that, all tokens are sent to the spatial encoder module to fully exploit the spatial sparse features of key points and model the spatial relationship.\\
However, just capturing the spatial relationship may affect the stability of the predicted pose sequence, so we add a temporal encoding module after the spatial encoding module to capture the context of the pose sequence over a long distance.Assuming that the length of the input pose sequence is T, the difference from the previous one is that we treat all key points of the body as a Token, and each Token contains J two-dimensional key points.After that, we perform temporal position encoding $E_T\in R^{N\times{T}}$ to record the positional relationship of the pose sequence in time.Finally, T number of tokens are fed into the time encoder to model the temporal relationship over long distances.
\subsection{Pose Trajectory Reconstruction}
The previous work \cite{pavllo20193d,zheng20213d} excavated temporal and spatial information on human pose sequences from a global perspective, but lacked in-depth analysis of local information.Recently, \cite{hassanin2022crossformer} have started to pay attention to the local features of joints and have achieved good performance.\cite{zhang2022mixste} models the trajectory of each keypoint in time by separating all keypoints.Although all the keypoints are transmitted simultaneously through the network, it will take up a lot of video memory resources.As shown in Figure \ref{fig4}(a), we adopt a relatively simple method to reconstruct the original pose sequence $X\in R^{T\times{J\times{2}}}$ into $\tilde{X}\in R^{J\times{T\times{2}}}$ through dimensional transformation.This process is called the reconstruction of the pose sequence, and the reconstructed features are called trajectory features.The final dimension of the reconstructed pose sequence is the trajectory corresponding to each key point.
\subsection{Local Interaction Module}
This section will elaborate on the details of the Local Interaction Module (LIM).As shown in Figure \ref{fig3}, the local information Interaction module includes two sub-modules: the self-trajectory encoder module (STE) and the cross-trajectory encoder module (CTE).To mine the temporal relationship of the same position key points, we consider each key point in the reconstructed pose trajectory sequence $\tilde{X}\in R^{J\times {T\times{D}}}$ as a token and input it into the self-trajectory encoder module (STE).Since there is also a corresponding relationship between different attitude trajectories, unlike before, we input the trajectory of each key point $x_j\in R^{T\times{D}}$ as a Token into the cross-trajectory encoder module (CTE) to model the deep correlation between different trajectories.
\subsection{Transformer Block}
In the global interaction module and the local interaction module, we use the transformer-based network structure.Attenton can be described as three matrices Q(query), K(key), V(value) obtained by dot product respectively, the specific formula is as follows:$$Attention(Q,K,V) = \text{softmax}(\frac{QK^T}{\sqrt{d_k}})V,$$where Q, K and V are essentially matrices obtained by linear transformation of the same input sequence, and self-attention is obtained through dot multiplication.Since a single head can only focus on a single angle of the feature, we can feed Q, K and V into n heads to obtain richer features, which are called Multi-Head Attention(MHA).The specific formula is as follows:$$MHA = f(Concat(Head_1,Head_2,\dots,Head_n)),$$$$\\{where}\;H_i=Attention(Q_i,K_i,V_i),i\in[1,\dots,n]$$where $f$ is the linear projection function.\\
In our method, the input of the self-trajectory encoder (STE) structure is each joint $p_j^\tau$ in a single trajectory, and the input of the cross-trajectory encoder (CTE) structure is $J$ trajectory sequence $P_j^T=\{ p_j^1,p_j^2,\dots,p_j^\tau,\dots,p_j^T\}$.Assuming that the encoded feature of the reconstructed sequence $\tilde{X}$ is $\tilde{X}_\phi\in R^{J\times{T\times{D}}}$, then the trajectory encoder can be represented as follows:$$\tilde{X}_\phi'=MSA(LN(\tilde{X}_\phi))+\tilde{X}_\phi,$$ $$\tilde{X}_\phi''=MLP(LN(\tilde{X}_\phi'))+\tilde{X}_\phi',$$ $$\\Y=LN(\tilde{X}_\phi''),$$where $LN(\cdot)$ represents layer normalization and $MLP(\cdot)$ represents multilayer perceptron.
\subsection{Regression Head}
Since our method adopts two parallel spatio-temporal encoding module and trajectory encoding module, we concatenate the spatio-temporal feature $X_\phi$ and the trajectory feature $\tilde{X}_\phi$ in the frame dimension to obtain $X_{concat }\in R^{2T\times({J\times{D}})}$.Since our method outputs all frames, we get the output $Y\in R^{T\times({J\times{D}})}$ by using 1D convolution on the frame dimension.Finally, we use a multi-layer perceptron $(MLP)$ consisting of a normalization layer and a linear layer for dimensionality reduction to obtain $Y\in R^{T\times({J\times{3}})}$.
\subsection{Pose Refinement}
Since the 2D extractor has poor extraction ability for some complex actions, its 3D projection has poor consistency.To address this issue, we use a pose refinement network (PR) with two linear layers to balance the effects of 2D pose.We feed two 3D pose sequences into the network (the first is the output predicted by the network, and the second is the connection between the true value of the 2D pose and the predicted depth of the first) to get the confidence of the two sets of results.Finally, the confidence is used to weight it as the final prediction result.

\subsection{Loss Function}
Our method builds a loss on all frames of the output and finally predicts intermediate frames.Our model employs a ${L2}$ loss to minimize the error between the predicted value and the true label:$$\mathcal{L}_m=\frac{1}{T}\frac{1}{J}\sum_{i=1}^{T}\sum_{j=1}^{J}||y_{i,j}-\tilde{y}_{i,j}||_2$$ where $y_{i,j}$ and $\tilde{y}_{i,j}$ represent the true label and predicted value of the $j$ joint on the $i$ frame, respectively.\\
Meanwhile, in order to improve the consistency of 3D projections, we introduce a pose fine-tuning loss $\mathcal{L}_r$ to fine-tune the predicted intermediate frames:$$\mathcal{L}_r=\frac{1}{J}\sum_{j=1}^{J}||y_{single}-\tilde{y}_{single}||_2$$\\
where $y_{single}$ and $\tilde{y}_{single}$ represent the true label and predicted value of the intermediate frame, respectively.\\
In our experiments, the total combined loss can be expressed as:$$\mathcal{L}=\lambda_m\mathcal{L}_m+\lambda_r\mathcal{L}_r$$
Where $\lambda_m$ and $\lambda_r$ correspond to the balance weights of $\mathcal{L}_m$ and fine-tune loss $\mathcal{L}_r$, respectively.
\newpage
\begin{table}[h]
\begin{center}

\caption{
Quantitative comparisons of Mean Per Joint Position Error (MPJPE) in millimeter between the estimated pose and the ground-truth
on Human3.6M under Protocol \#1 and Protocol \#2, where f denotes the number of input frames used in each method. The best score is
marked in bold.
}
\label{lab1}
\resizebox{16.0cm}{3.6cm}{
\begin{tabular}{l|lcccccccccccccc|l}
\toprule[2pt]
\noalign{\smallskip}

Protocol {\#1} $\qquad\qquad$	                      &Dir.&Disc.&Eat &Greet &Phone &Photo &Pose &Pur. &Sit &SitD. &Smoke &Wait &WalkD. &Walk &WalkT & Avg\\
\noalign{\smallskip}
\hline
\noalign{\smallskip}
Tekin {\it et al.}\cite{tekin2017learning}ICCV’17       &54.2 &61.4 &60.2 &61.2 &79.4 &78.3 &63.1 &81.6 &70.1 &107.3 &69.3 &70.3 &74.3 &51.8 &63.2&69.7\\
Martinez {\it et al.}\cite{Sun_2017_ICCV}ICCV’17        &51.8 &56.2 &58.1 &59.0 &69.5 &78.4 &55.2 &58.1 &74.0 &94.6 &62.3 &59.1 &65.1 &49.5 &52.4 &62.9\\
Fang {\it et al.}\cite{fang2018learning}AAAI’18         &50.1 &54.3 &57.0 &57.1 &66.6 &73.3 &53.4 &55.7 &72.8 &88.6 &60.3 &57.7 &62.7 &47.5 &50.6 &60.4\\
Hossain {\it et al.}\cite{hossain2018exploiting}ECCV’18 &48.4 &50.7 &57.2 &55.2 &63.1 &72.6 &53.0 &51.7 &66.1 &80.9 &59.0 &57.3 &62.4 &46.6 &49.6 &58.3\\
Pavlakos {\it et al.}\cite{pavlakos2018ordinal}CVPR'18   &48.5 &54.4 &54.4 &52.0 &59.4 &65.3 &49.9 &52.9 &65.8 &71.1 &56.6 &52.9 &60.9 &44.7 &47.8 &56.2\\
Lee {\it et al.}\cite{lee2018propagating}ECCV'18&\textbf{40.2} &\textbf{49.2} &47.8 &52.6 &\textbf{50.1} &75.0 &50.2 &\textbf{43.0} &\textbf{55.8} &73.9 &54.1 &55.6 &58.2 &43.3 &43.3 &52.8\\
Pavllo {\it et al.}\cite{pavllo20193d}arxiv’18          &47.1 &50.6 &49.0 &51.8 &53.6 &61.4 &49.4 &47.4 &59.3 &67.4 &52.4 &49.5 &55.3 &39.5 &42.7 &51.8\\
GraphSH \cite{xu2021graph}CVPR’21                       &45.2 &49.9 &47.5 &50.9 &54.9 &66.1 &48.5 &46.3 &59.7 &71.5 &51.4 &48.6 &53.9 &39.9 &44.1 &51.9\\
PoseFormer \cite{zheng20213d} ICCV’21(f=9) &-&-&-&-&-&-&-&-&-&-&-&-&-&-&-&49.9\\
MGCN \cite{zou2021modulated}ICCV’21                     &45.4 &\textbf{49.2} &\textbf{45.7} &49.4 &50.4 &58.2 &47.9 &46.0 &57.5 &\textbf{63.0} &49.7 &46.6 &52.2 &38.9 &40.8 &49.4\\
\hline
\noalign{\smallskip}
ours(f=9)                                                &44.0 &49.4 &46.4 &\textbf{47.3} &51.0 &\textbf{58.0} &\textbf{45.5} &45.2 &58.9 &66.5 &\textbf{49.1} &\textbf{45.8} &\textbf{50.8} &\textbf{35.2} &\textbf{38.0} &\textbf{48.7}\\
\noalign{\smallskip}
\toprule[2pt]
\noalign{\smallskip}
Protocol {\#2} $\qquad\qquad$	                      &Dir.&Disc.&Eat &Greet &Phone &Photo &Pose &Pur. &Sit &SitD. &Smoke &Wait &WalkD. &Walk &WalkT & Avg\\
\noalign{\smallskip}
\hline
\noalign{\smallskip}
Martinez {\it et al.}\cite{Sun_2017_ICCV}ICCV’17     &39.5 &43.2 &46.4 &47.0 &51.0 &56.0 &41.4 &40.6 &56.5 &69.4 &49.2 &45.0 &49.5 &38.0 &43.1 &47.7\\
Sun {\it et al.}\cite{liang2018compositional}ICCV'17  &42.1 &44.3 &45.0 &45.4 &51.5 &53.0 &43.2 &41.3 &59.3 &73.3 &51.0 &44.0 &48.0 &38.3 &44.8 &48.3\\
Fang {\it et al.}\cite{Fang2018LearningPG}AAAI’18    &38.2 &41.7 &43.7 &44.9 &48.5 &55.3 &40.2 &38.2 &54.5 &64.4 &47.2 &44.3 &47.3 &36.7 &41.7 &45.7\\
Pavlakos {\it et al.}\cite{pavlakos2018ordinal}CVPR'18&34.7 &39.8 &41.8 &38.6 &42.5 &47.5 &38.0 &36.6 &50.7 &56.8 &42.6 &39.6 &43.9 &32.1 &36.5 &41.8\\
Hossain {\it et al.}\cite{hossain2018exploiting}ECCV’18 &35.7 &39.3 &44.6 &43.0 &47.2 &54.0 &38.3 &37.5 &51.6 &61.3 &46.5 &41.4 &47.3 &34.2 &39.4 &44.1\\
Lee {\it et al.}\cite{lee2018propagating}ECCV'18      &34.9 &\textbf{35.2} &43.2 &42.6 &46.2 &55.0 &37.6 &38.8 &50.9 &67.3 &48.9 &35.2 &50.7 &31.0 &34.6 &43.4\\
Pavllo {\it et al.}\cite{pavllo20193d}arxiv’18       &36.0 &38.7 &38.0 &41.7 &40.1 &45.9 &37.1 &35.4 &46.8 &53.4 &41.4 &36.9 &43.1 &30.3 &34.8 &40.0\\
cai {\it et al.}\cite{Cai_2019_ICCV}ICCV’19	              &35.7 &37.8 &36.9 &40.7 &39.6 &45.2 &37.4 &\textbf{34.5} &46.9 &\textbf{50.1} &40.5 &36.1 &41.0 &29.6 &33.2 &39.0\\
\hline
\noalign{\smallskip}
ours(f=9)                                            &\textbf{33.9} &38.0 &\textbf{36.4} &\textbf{38.0} &\textbf{38.4} &\textbf{44.3} &\textbf{34.0} &34.9 &\textbf{46.7} &51.9 &\textbf{39.1} &\textbf{34.4} &\textbf{40.0} &\textbf{27.5} &\textbf{31.1} &\textbf{37.9}\\
\noalign{\smallskip}
\toprule[2pt]

\end{tabular}}
\end{center}
\end{table}

\begin{table}[h]
\begin{center}
\caption{
Quantitative comparison with the state-of-the-art methods
on MPI-INF-3DHP.The best score is marked in bold.\\
}
\label{lab2}
\resizebox{10cm}{2.2cm}{
\begin{tabular}{l|ccc}
\toprule[2pt]
\noalign{\smallskip}
Method &PCK$\uparrow$ &AUC$\uparrow$ &MPJPE$\downarrow$  \\
\noalign{\smallskip}
\hline
\noalign{\smallskip} 
Mehta {\it et al.}\cite{DBLP:journals/corr/MehtaRCSXT16} 3DV’17 (f=1) &75.7 &39.3 &117.6\\
Pavllo {\it et al.}\cite{pavllo20193d} CVPR’19(f=81) &86.0 &1.9  &84.0\\
Lin {\it et al.}\cite{Lin2019TrajectorySF}BMVC’19(f=25) &83.6 &51.4 &79.8\\
wang {\it et al.}\cite{DBLP:journals/corr/abs-2004-13985}ECCV’20(f=96)  &86.9 &62.1 &68.1\\
zheng {\it et al.}\cite{zheng20213d}ICCV’21(f=9)  &88.6 &56.4 &77.1\\
zhang {\it et al.}\cite{zhang2022mixste} CVPR' 22(f=27) &94.4 &66.5 &54.9\\
Hu {\it et al.}\cite{2021Conditional}MM’21 (f=96) &97.9 &69.5 &42.5\\
\noalign{\smallskip}
\hline
\noalign{\smallskip} 
ours(f=9) &\textbf{97.9} &\textbf{70.0} &\textbf{28.2}\\
\noalign{\smallskip}
\toprule[2pt]
\end{tabular}}
\end{center}
\end{table}

\noindent			
\section{Experiments}
\subsection{Datasets and Evaluation Metrics}
\textbf{Human3.6M} is the most commonly used and largest indoor dataset for HPE at this stage.This dataset mainly captures the 3D pose of the human body from four perspectives.This dataset mainly captures the 3D pose of the human body from four perspectives. These poses consist of 15 actions such as eating, sitting, walking, and making phone calls.It contains 3.6 million video frame pictures.The pictures are divided into 11 parts, however we use five parts (S1, S5, S6, S7, S8) as training set and two parts (S9 and S11) as test set.We adopt MPJPE as the evaluation metric on this dataset, which is defined as the Euclidean distance between the predicted value and the true value.\\
\textbf{MPI-INF-3DHP} is a 3D human pose estimation dataset consisting of indoor and outdoor scenes.The dataset consists of more than 1.3 million frames of images, recording 8 categories of activities of 8 participants from 14 camera angles.For the MPI-INF-3DHP dataset, we adopt MPJPE, PCK (Percentage of Correct Keypoint) and AUC (Area Under Curve) as evaluation metrics according to previous methods.	
\subsection{Implementation Details}		
In our experiments, we used two 2080Ti NVIDIA GPUs and used the current mainstream pytorch framework.When training the model, we used the Adam optimizer.We set the initial learning rate to 0.001 and multiply it by a jitter factor of 0.95 after every fifth iteration, for a total of 40 iterations.According to \cite{zheng20213d}, our method Fusionformer uses 4 layers of spatial encoder (SE)and 4 layers of temporal encoder (TE) to better learn spatio-temporal features.Following the previous method \cite{cai2019exploiting,chen2021anatomy,pavllo20193d,sun2018integral}, we also adopt a flip strategy for the input to augment the data.We adopt CPN extractor for Human3.6M dataset and 2D ground truth labels for MPI-INF-3DHP dataset respectively.\\		
\subsection{Comparison with State-of-the-Art Methods}
\textbf{Results on Human3.6M.}Our proposed Fusionformer is compared with previous mainstream methods on the Human3.6M dataset.Table \ref{lab1} shows the results of our model with an input of 9 frames.To our surprise, our method outperforms the previous state-of-the-art methods under two metrics, Protocol $\#1$ (48.7mm) and Protocol $\#2$ (37.9mm).Compared with the recent GCN-based network (MGCN) and Transformer-based network (PoseFormer), our method is improved by 0.7mm and 1.2mm on MPJPE, respectively.\\\\
\textbf{Results on MPI-INF-3DHP.}To further examine the generalizability of the model, we also evaluated our method on the MPI-INF-3DHP dataset.In order to test the performance of the model in the low input frame scene, we also use the 9-frame 2D pose sequence as the input of the model.Compared to previous state-of-the-art methods, our method even outperforms some methods in multi-input frame scenarios.Table \ref{lab2} demonstrates that our method leads the way on all metrics (PCK, AUC, MPJPE).At the same time, this also reflects the adaptability of our model to indoor and outdoor and other scenarios.\\\\
\subsection{Ablation Study}
\textbf{Impact of Receptive Fields.}For video-based 3D human pose estimation tasks, low input video frames are of great significance for improving the output speed and efficiency of the network. We set up four sets of comparative experiments under different input frames in our experiments.Table \ref{lab3} shows the results under different input frames. From the results, it can be seen that using the CPN extractor as the input reduces the error rate by 5\% from 3 to 9 frames, and using the GT as the input obtains more gain.The results show that our method can still achieve good performance at low input frames.In the following part of the ablation experiment, we all use the input scene of 9 frames.\\\\\\
\textbf{Effect of Model Components.}In Table \ref{lab4}, we analyze the impact of each module on network performance through experiments.Firstly, we compare our approach with baseline(poseformer).For a fair comparison, we set the same number of layers and embedding dimensions in both spatio-temporal modules.From the experimental results, it can be found that we improve MPJPE by 0.5mm after introducing the self-trajectory module (STE) and cross-trajectory (CTE) module.At the same time, we found that the self-trajectory module has a greater performance improvement on the network than the trajectory module.Finally, we introduce a fine-tuning network with an overall performance improvement of 0.7mmMPJPE.These ablation experiments show that our Fusionformer can not only explore the temporal relationship of joints well, but also can better coordinate the motion of different joints.Fine-tuning the network has played a great role in improving the consistency of 3D projections.

\begin{minipage}{\textwidth}
        \begin{minipage}[t]{0.3\textwidth}
            \centering
            \makeatletter\def\@captype{table}\makeatother\caption{Ablation study on different receptive fields with MPJPE(mm). CPN - cascaded pyramid network; GT - 2D ground truth.\\}
            \begin{tabular}{ccccc}

\toprule[2pt]
\noalign{\smallskip}
 &3 &5 &7 &9\\
\noalign{\smallskip}
\hline
\noalign{\smallskip} 
CPN &51.5 &50.1 &49.2 &48.7\\
\noalign{\smallskip}
\noalign{\smallskip} 
GT &41.7 &40.8 &40.6 &36.8\\
\noalign{\smallskip}
\toprule[2pt]
\label{lab3}
\end{tabular}
        \end{minipage}
        \begin{minipage}[t]{0.7\textwidth}
        \centering
        \makeatletter\def\@captype{table}\makeatother\caption{\label{lab4} Ablation study on different components of our Fusionormer. Three models(STE,CTE,PR) are tested in the table below.\\}
            \begin{tabular}{l|ccc|c}

\toprule[2pt]
\noalign{\smallskip}
Method &STE &CTE &PR &MPJPE(mm)\\
\noalign{\smallskip}
\hline
\noalign{\smallskip}
Baseline &\XSolidBrush &\XSolidBrush &\XSolidBrush &49.9\\
STE-PR &\Checkmark &\XSolidBrush &\Checkmark &49.3\\
CTE-PR &\XSolidBrush &\Checkmark &\Checkmark &49.7\\
CTE-PR &\Checkmark  &\Checkmark &\XSolidBrush &49.4\\
Fusionformer(ours) &\Checkmark  &\Checkmark &\Checkmark &48.7\\
\toprule[2pt]

\end{tabular}
        \end{minipage}
    \end{minipage}
\textbf{Impact of Parameters in STE and CTE.}Table \ref{lab5} shows how different parameter settings of STE and CTE affect the performance and computational complexity of the model.The results show that we get poorer performance with lower number of layers for STE and CTE, but gain consistently by stacking layers continuously.After setting $L_s=4$ and $L_c=4$ respectively, we observe that no gain is obtained anymore.Due to the shallower network stacking, it can be seen from the table that our parameter amount and computational complexity are low.It should be noted here that our ablation experiments on the parameter of embedding dimension indicate that there was almost no performance improvement.

\begin{table}[h]
\begin{center}
\caption{\label{lab5}Ablation study on different parameters of STE and CTE.Here, $L_s$ and $L_c$ indicate the number of STE and CTE layers,respectively.\\}
\begin{tabular}{cccccc}
\toprule[2pt]
\noalign{\smallskip}
&$L_s$ &$L_c$ &$Params (M)$ &$FLOPs (G)$ &$MPJPE$\\
\noalign{\smallskip}
\hline
\noalign{\smallskip} 
&2 &2 &10.91&0.46 &49.9\\
\noalign{\smallskip}
&2 &3 &11.58&0.51 &49.1\\
\noalign{\smallskip}
&3 &2 &10.92&0.47 &49.2\\
\noalign{\smallskip}
&3 &3 &11.59&0.51 &48.9\\
\noalign{\smallskip}
&4 &4 &12.26&0.56&48.7\\
\noalign{\smallskip}
&5 &5 &12.92&0.61&49.0\\
\noalign{\smallskip}
&6 &6 &13.59&0.66&49.3\\
\noalign{\smallskip}
\toprule[2pt]
\end{tabular}
\end{center}
\end{table}

\subsection{Qualitative Results}
\begin{figure}[h]
\centering
\includegraphics[width=0.6\textwidth]{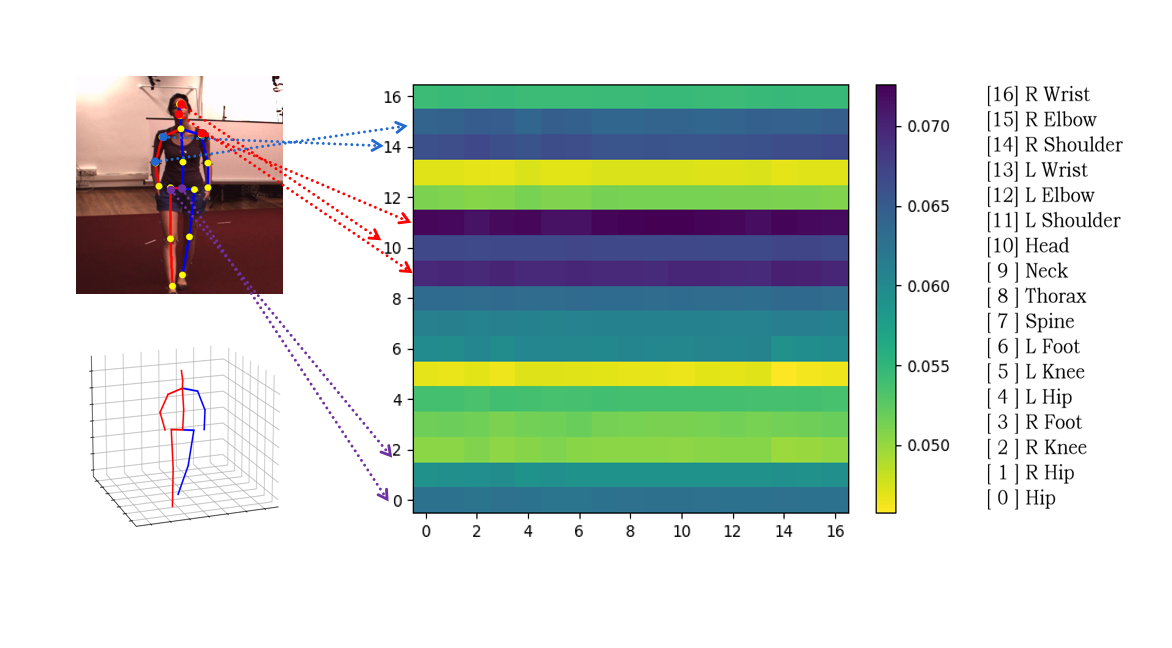}
\caption{Visualization of self-attentions among different joint trajectories. The x-axis and y-axis correspond to the queries and the predicted outputs, respectively. Each row shows the attention weight wi,j of the j-th query for the i-th output.}
\label{fig5}
\end{figure}

\begin{figure}[h]
\centering
\includegraphics[width=0.65\textwidth]{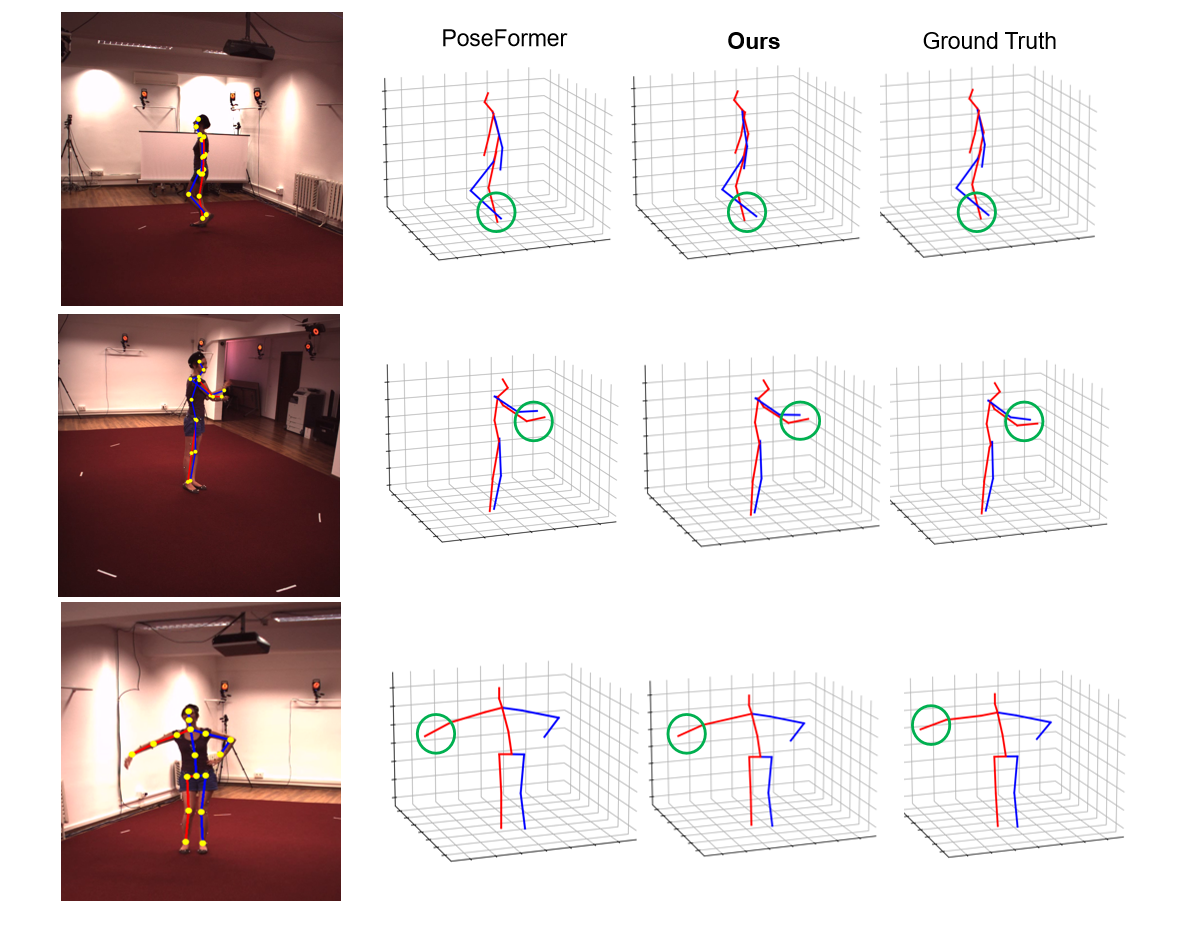}
\caption{Qualitative comparison among the proposed method (Fusionformer),
PoseFormer on Human3.6M dataset.}
\label{fig6}
\end{figure}

From Figure \ref{fig5}, in order to show the synergy between joints, we further visualize the trajectory attention of different joints.We selected 9 consecutive frames under the walking action for attention visualization.From the trajectory attention map, we can observe the time dependence of different joints.Some adjacent joints have higher and similar attention weights in time, which represents the synergy of these joints in motion.Moreover, Figure \ref{fig6} shows the visualization results of our method and PoseFormer in low input frame scenarios. We selected three different motions from multiple viewpoints to capture. The results show that our model can more accurately project the 3D pose of the human body under certain complex actions.


\section{Conclusion}
In this paper, we propose Fusionformer, a Transformer-based fusion network.Similar to the two-stream network, we introduce two branches GIM and LIM.The GIM module handles the global spatio-temporal correlation of human poses.LIM not only models the temporal trend of a single joint, but also deeply mines the temporal synergy of different joints.Moreover,we use a local-global approach to fuse the two features.Finally, we also introduce a fine-tuning network to eliminate the influence of poor 2D pose on 3D projection consistency to further improve the stability of the model.The results two benchmark datasets (Human3.6M, MPI-INF-3DHP) demonstrate the excellent performance of our model on the task of 3D human pose estimation.\\

\bibliographystyle{plain}
\bibliography{sample}

\end{document}